# AI-Based Demand Forecasting and Load Balancing for Optimising Energy use in Healthcare Systems: A real case study

Iman Rahimi*, Isha Patel
Faculty of Engineering & Information Technology, University of Technology Sydney, Australia (*corresponding author: iman83@gmail.com)

**Abstract-** This paper addresses the critical need for efficient energy management in healthcare facilities, where fluctuating energy demands challenge both operational and sustainability goals. Traditional energy management methods often fall short in healthcare settings, leading to inefficiencies and increased costs. To address this, the paper explores AI-driven approaches for demand forecasting and load balancing, introducing a novel integration of LSTM (Long Short-Term Memory), genetic algorithm, and SHAP (Shapley Additive Explanations) specifically tailored for healthcare energy management. While LSTM has been widely used for time-series forecasting, its application in healthcare energy demand prediction is underexplored. Here, LSTM is demonstrated to significantly outperform ARIMA and Prophet models in handling complex, non-linear demand patterns. Results show that LSTM achieved a Mean Absolute Error (MAE) of 21.69 and Root Mean Square Error (RMSE) of 29.96, significantly improving upon Prophet (MAE: 59.78, RMSE: 81.22) and ARIMA (MAE: 87.73, RMSE: 125.22), highlighting its superior forecasting capability. Genetic algorithm is employed not only for optimising forecasting model parameters but also for dynamically improving load balancing strategies, ensuring adaptability to real-time energy fluctuations. Additionally, SHAP analysis is used to interpret the models and understand the impact of various input features on predictions, enhancing model transparency and trustworthiness in energy decision-making. The combined LSTM-GA-SHAP approach offers a comprehensive framework that improves forecasting accuracy, enhances energy efficiency, and supports sustainability in healthcare environments. Future work could focus on real-time implementation and further hybridisation with reinforcement learning for continuous optimisation. This study establishes a strong foundation for leveraging AI in healthcare energy management, showcasing its potential for scalability, efficiency, and resilience.

**Keywords:**
Healthcare energy optimisation, demand forecasting, Genetic Algorithm, renewable energy integration, sustainability.

# 1. Introduction

Australia has a big capacity of using renewable energy in different regions (Holloway, R, 2023; Rahimi et al., 2025). Australian healthcare system plays a major role in using renewable energies. Optimising energy use in healthcare systems is essential due to the high and often unpredictable energy demands needed to run medical equipment, keep environmental conditions stable, and support constant patient care.
Traditional energy management methods often don't meet the complex needs of healthcare environments, where energy usage patterns can vary based on patient intake, equipment use, and seasonal factors (Prananda, Dini, & Putri, 2021).
AI based demand forecasting offers a valuable solution by using past data, operational activities, and environmental conditions to accurately predict future energy needs (Masri, Zeineldin, & Woon, 2015). Machine learning algorithms, like neural networks (NN), support vector machines (SVMs), and time series analysis, can pick up on complex relationships between different factors that influence energy demand. For example, LSTM neural networks are particularly effective for managing sequential data, which improves short- and medium-term energy forecasts by learning from trends over time. This makes them especially useful in healthcare, where demand can shift frequently (Doshi, Mridha, Kumar, & Ved, 2021).

By using AI-powered demand forecasting, healthcare facilities can better prepare for high demand periods, allocate resources more efficiently and keep energy costs down (Rani, Kumar, & Dhingra, 2022). Hybrid forecasting approaches that combine machine learning models like LSTM with traditional methods offer even more accuracy. These are especially helpful in healthcare settings with intermittent or unpredictable energy needs, as hybrid models allow systems to handle complex demand patterns more flexible and avoid both under and overuse of resources (S & S, 2022).

To further interpret the forecasting models and gain insights into their decision-making processes, SHAP (Shapley Additive Explanations) analysis is employed. SHAP values help explain the contribution of each feature in the prediction of energy demand by quantifying the impact of different inputs, such as historical energy usage, environmental conditions, and operational factors (Antwarg, Miller, Shapira, & Rokach, 2021). By applying SHAP analysis to the ARIMA, Prophet, and LSTM models, this study offers a clearer understanding of how each model interprets the data, allowing for more transparent and interpretable forecasting. This method not only highlights the importance of specific features but also reveals potential biases in the models, making it a valuable tool for enhancing the accuracy and reliability of energy demand predictions in healthcare environments.

Load balancing is another essential part of energy management in healthcare. It distributes energy loads across departments and different time frames, preventing overloads and making resource use more efficient (Patni & Aswal, Distributed load balancing model for grid computing environments, 2015). Dynamic load balancing algorithms, inspired by grid computing, reallocate energy based on real time usage, forecasted needs, and renewable energy availability. Adaptive load balancing techniques, such as those used in Software Defined Networking (SDN), could inform healthcare load balancing by automatically redistributing energy to prevent any single department from being matched to current demand, avoiding power disruptions and ensuring the system runs smoothly (Lan, Li, Liu, & Qiu, 2017).

For resilient and flexible energy management, AI driven hierarchical load balancing models help optimise distribution across local, group, and network levels. Adapted from distributed grid computing models, these systems allow healthcare facilities to scale energy allocation according to their size and needs, lowering communication costs and response times while maximising energy use. By organising load balancing hierarchically, healthcare systems meet energy demands at local (within a facility), group (across facilities), and network levels efficiently. This layered approach also decreases the likelihood of widespread disruptions, as energy loads are adjusted in real time across different levels (Patni & Aswal, Dynamic load balancing model for layered grid architecture, 2015).

By combining AI based demand forecasting with adaptive load balancing, healthcare facilities gain a complete energy management system that improves both operational efficiency and sustainability. Genetic algorithm (GA) and Reinforcement Learning (RL) give the system adaptive abilities, enabling it to respond to fluctuations in demand in real time. These AI driven strategies enhance operational reliability while supporting environmental goals by making better use of renewable energy sources and reducing reliance on traditional energy supplies. Aligning AI driven forecasting with load balancing allows healthcare facilities to maintain energy stability, cut costs, and work toward sustainability. This approach highlights how AI advancements are providing healthcare with scalable, adaptable, and environmentally friendly energy solutions (Hameed, Hashemi, & Træholt, 2021).

This study contributes to the literature by developing an integrated AI-driven framework that merges demand forecasting techniques, including ARIMA, Prophet, and LSTM models, with Genetic Algorithm-based load balancing, aiming to optimize energy usage within healthcare facilities. Furthermore, the research utilizes SHAP analysis to enhance model interpretability, thereby offering valuable insights into the influence of key variables on energy demand predictions. It evaluates the performance of different forecasting models in conjunction with GA for load balancing, providing a comprehensive comparative analysis to identify the most efficient strategy. Additionally, the study introduces a hierarchical load balancing approach designed to enhance energy distribution efficiency and resilience across various levels of healthcare networks. In addressing existing research gaps, this work uniquely integrates AI-based forecasting with adaptive load balancing in a cohesive framework specifically tailored for healthcare energy management.

This research aims to explore how different demand forecasting models – ARIMA, Prophet, and LSTM – combined with a GA, can help manage energy use in healthcare facilities. The study's goal is to identify the most effective model combination for predicting energy demand and balancing energy load efficiently, supporting cost savings and operational stability in healthcare settings. At the end, the following research questions are answered:
1. How well do ARIMA, Prophet, and LSTM models perform in predicting energy demand for healthcare facilities?
2. How does integrating GA with each forecasting model affect energy load balancing within healthcare settings?
3. Which forecasting model combined with GA offers the most practical balance of prediction accuracy, load balancing efficiency, and cost savings?
4. How can AI-driven energy management systems be effectively implemented in healthcare facilities for better resource use and sustainability?

The rest of the paper is organised as follows. Section 2 presents the literature review, which highlights previous studies on demand forecasting and load balancing in energy management, showing where current research could improve. The methodology section explains the machine learning models, and optimisation techniques used to forecast energy demand and balance loads effectively. The results analysis section evaluates the performance of ARIMA, Prophet, and LSTM models, demonstrating LSTM's superior predictive accuracy and the role of GA in optimising load balancing. The discussion interprets these findings in relation to the study's aims, emphasising the impact of AI-driven forecasting on energy efficiency and cost reduction in healthcare settings. Finally, the conclusion and future work sections summarise the study's contributions, highlighting the effectiveness of the proposed LSTM-GA framework and suggesting areas for further research, including real-time implementation and hybrid modelling approaches.

## 2. Literature Review

Healthcare facilities have high energy demands due to continuous patient care, medical equipment operation, and strict environmental controls. As energy costs rise and sustainability becomes more critical, healthcare facilities need smarter ways to manage energy (Benefits of Energy Efficiency in Healthcare, 2023).

Demand forecasting involves predicting future energy needs in healthcare facilities by analyzing historical data patterns (Huang, Xu, Ji, Xiang, & He, 2020). Common forecasting methods include statistical models such as Autoregressive Integrated Moving Average (ARIMA), machine learning approaches like Long Short-Term Memory (LSTM) neural networks, and specialized time series tools like Prophet. Load balancing, on the other hand, refers to the distribution of energy usage across different departments or time periods to prevent overloads and maximize efficiency. GA are frequently employed for this purpose, as they iteratively search for optimal resource allocation by simulating natural selection processes (Singhal, Sharma, Anushree, & Verma, 2024). Additionally, AI-driven optimization leverages machine learning and GA to allocate energy resources effectively based on real-time forecasts, enabling facilities to adjust to demand fluctuations, enhance energy utilization, and reduce costs (Antonopoulos, et al., 2020).

Energy management in healthcare is complex because it must support lifesaving equipment and constant patient care without interruption. Initially, hospitals managed energy through manual adjustments and traditional statistical models, but these approaches couldn't handle the complexity of demand patterns in healthcare (Junaid, et al., 2022). With advancements in AI, researchers started using machine learning models for demand forecasting and optimisation algorithms for load balancing, offering a more efficient solution. Historically, ARIMA and other traditional time series models were the go-to for demand forecasting, as they're straightforward and perform well with linear data (Bajaj , 2023). However, neural networks like LSTM have gained attention for their ability to capture long term patterns and dependencies, making them suitable for more complex, non-linear demand trends in healthcare. Similarly, GA were initially used in manufacturing and logistics for resource allocation, but their flexibility and adaptability make them well suited to managing energy needs in dynamic environments like healthcare (Bolstad, Cali, Kuzlu, & Halden, 2022).

Machine learning models, particularly neural networks, have demonstrated superior performance in energy forecasting compared to traditional models, as highlighted by Masri et al. (2015). Their study underscores how advanced algorithms can effectively predict energy demand patterns in dynamic environments. Similarly, Doshi et al. (2021) emphasized that Long Short-Term Memory (LSTM) networks are particularly effective in capturing healthcare-specific demand fluctuations, achieving higher accuracy than simpler forecasting models by identifying complex temporal dependencies.

Load balancing models have also gained prominence in energy management contexts. Patni and Aswal (2015) explored distributed load balancing in grid computing, which has since inspired adaptations for energy management in healthcare. Additionally, Lan et al. (2017) examined the application of adaptive load balancing, as seen in Software Defined Networking (SDN), to real-time energy distributions. Their findings suggest that dynamically adjusting energy allocations based on current demand can significantly enhance system efficiency and resilience.

GA have proven particularly effective in optimizing energy resource management, given their capacity to handle complex and dynamic requirements. Hameed et al. (2021) demonstrated that GA's adaptability to real-time adjustments allows for effective load balancing, making them particularly suitable for unpredictable energy demands in healthcare settings. Their study confirms that GA can optimize energy distribution effectively, accounting for both demand fluctuations and operational constraints.

As summarised in Table 1, traditional models like ARIMA perform well for stationary and linear data but struggle with capturing complex, non-linear demand patterns commonly found in healthcare energy usage. Prophet improves upon ARIMA by handling missing data and seasonal variations more effectively but remains limited in highly irregular demand scenarios. LSTM, demonstrated superior accuracy in modelling long-term dependencies and non linear trends, however its higher computational costs and need for extensive parameter tuning highlight the importance of optimisation techniques like GA to enhance efficiency and ensure practical deployment in real world healthcare settings.

*Table1, Comparison of Previous Studies v This study*

| Study | Model | Key Findings | Limitations | Improvements in this study |
| --- | --- | --- | --- | --- |
| **(Masri, Zeineldin, & Woon, 2015)** | Neural Networks | Machine Learning models outperform traditional forecasting models | Lacked optimisation for real time load balancing | Integrated GA for dynamic load balancing |
| **(Doshi, Mridha, Kumar, & Ved, 2021)** | LSTM | LSTM effectively captures healthcare energy demand fluctuations | Lacks explainability and real time adaptability | Used SHAP for interpretability and GA for optimisation |
| **(Patni & Aswal, Distributed load balancing model for grid computing** | Grid-based load balancing | Load balancing models improve resource distribution | Did not incorporate forecasting for predictive energy allocation | Incorporated LSTM forecasting for proactive energy balancing |
| environments, 2015) | | | | |
| (Lan, Li, Liu, & Qiu, 2017) | Adaptive load balancing | Real time load balancing enhances energy efficiency | Did not use demand forecasting to predict the optimise loads | Combined LSTM forecasting with GA for demand aware load balancing |
| (Hameed, Hashemi, & Træholt, 2021) | GA based load balancing | GA effectively optimises energy distribution | Did not integrate forecasting for proactive energy allocation | Used LSTM-GA integration for predictive and adaptive energy management |
| This Study | LSTM, ARIMA, Prophet, and GA with SHAP interpretability | Combines forecasting, optimisation, and interpretability for a comprehensive AI driven energy management approach | Addresses previous gaps in real time adaptability, forecasting accuracy, and explainability | Provided an integrated, scalable, and interpretable solution tailored for healthcare energy management |

The currentThis study combines ideas from time series forecasting and evolutionary algorithms for an integrated approach. Time Series Forecasting is essential in predicting future data points based on historical data. Table 2 provides forecasting model comparison used in the current study:For example, Table 12:

*Table 2, Forecasting Model Comparisons*

| Model | Description | Strengths | Limitations |
|---|---|---|---|
| **ARIMA** | A statistical model used for analysing and forecasting time series data, effective for linear trends (Bajaj , 2023). | Works well with stationary and linear data.<br><br>Easy to interpret and implement.<br><br>Requires minimal computational power. | Struggles with non-linear and complex demand patterns.<br><br>Requires data preprocessing (stationarity and differencing) |
| **Prophet** | A forecasting model developed by Facebook, designed to handle | Handles missing data well. | May struggle with highly irregular and |

|  | seasonality and missing data, suitable for demand patterns that vary seasonally (Python API, n.d.). | Captures seasonality and trend changes automatically. Requires minimal parameter tuning. | non-seasonal patterns. Less effective for very short-term forecasting |
| --- | --- | --- | --- |
| **LSTM** | A deep learning model designed to recognise patterns in sequential data, making it useful for complex time series forecasting (Tam, 2023). | Captures long term dependencies. Handles complex, non-linear relationships. Adaptable to large and irregular datasets. | High computational cost. Requires large datasets and tuning. Less interpretable than traditional models. |

For load balancing, GA provides a structure where various potential solutions (i.e., ways to allocate energy) evolved over iterations to find the most optimal distribution based on forecasted demands.

In recent years, there has been a shift toward combining multiple forecasting and optimisation techniques for a hybrid approach to energy management. Many studies pair traditional statistical models with machine learning to capture both linear and non-linear patterns in demand (Aghaabbasi & Chalermpang, 2023). However, while much research has focused on improving forecasting accuracy, fewer studies have examined how well different forecasting models perform when paired with optimisation algorithms like GA in healthcare. This study will address these gaps by testing three forecasting models – ARIMA, Prophet, and LSTM – with GA to find the best performing combination for balancing energy loads in healthcare settings.

Key debates include whether complex machine learning models are worth the extra computing resources compared to traditional methods. Some researchers argue that models like LSTM are highly accurate but can be overkill due to the computational demands and lack of interpretability (Prananda, Dini, & Putri, 2021). Similarly, GA and other optimisation techniques can be complex to implement, leading some to advocate for simpler, rule-based approaches in healthcare.

The current study will examine how AI based forecasting and load balancing methods can improve energy efficiency in healthcare. A literature review here helps establish the concepts, past research, and current developments in this area, providing a foundation for identifying research gaps and areas for improvement.
Building on this existing research, this study evaluated three demand forecasting models (ARIMA, Prophet, LSTM) paired with a GA to balance energy loads effectively in healthcare. By comparing the accuracy and efficiency of each combination, this research will help determine which forecasting model is most practical and effective when used alongside GA for energy management.

## 3. Research Methodology

The flowchart, Figure 1, presents a structured framework for AI-driven energy management in healthcare facilities, comprising five key components. The process begins with Data Collection & Preprocessing, which serves as the input for three forecasting models: ARIMA, Prophet, and LSTM, each designed to predict energy demand based on historical data. The forecasting outputs are then fed into the GA Optimization Parameter Tuning & Load Balancing module, which optimizes load allocation across multiple levels (Local, Group, Network) using GAs. The SHAP Analysis module interprets the forecasting model outputs, providing insights into feature importance and model behavior. The framework proceeds to the Evaluation & Monitoring phase, where key metrics such as MAE, RMSE, energy savings, and cost reduction are assessed to validate model performance. Finally, the Future Implementation stage outlines potential enhancements, including hybrid model integration and real-time deployment, ensuring scalability and adaptability in dynamic healthcare settings.

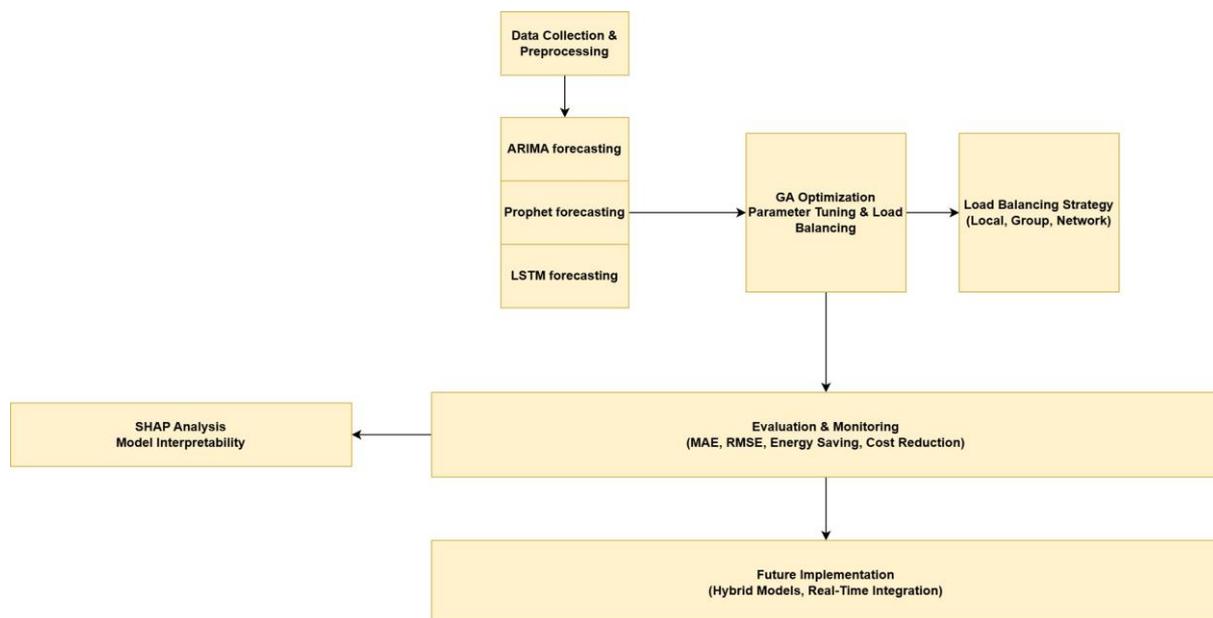

*Figure 1 a structured framework for AI-driven energy management in healthcare facilities*

### 3.1 Data Exploration

This study analyses energy consumption data from a hospital in Perth to understand usage patterns and identify potential efficiency improvements. Perth, shown in Figure 2, was selected as the focus of this study due to its unique climate, energy pricing structure, and healthcare infrastructure challenges. Unlike cities on the east coast, Perth experiences hot, dry summers and mild winters, creating distinct heating and cooling demands in healthcare facilities (Tong

& Wondmagegn, 2021). Additionally, Western Australia operates on a separate energy grid (the SWIS), which makes energy optimisation particularly important to ensure resilience and cost efficiency (Australian Energy Market Operator, 2021). Hospitals are among the highest energy consuming public facilities, and studying one in Perth offers insights into how building operations can be improved in similarly isolated or resource sensitive environments.

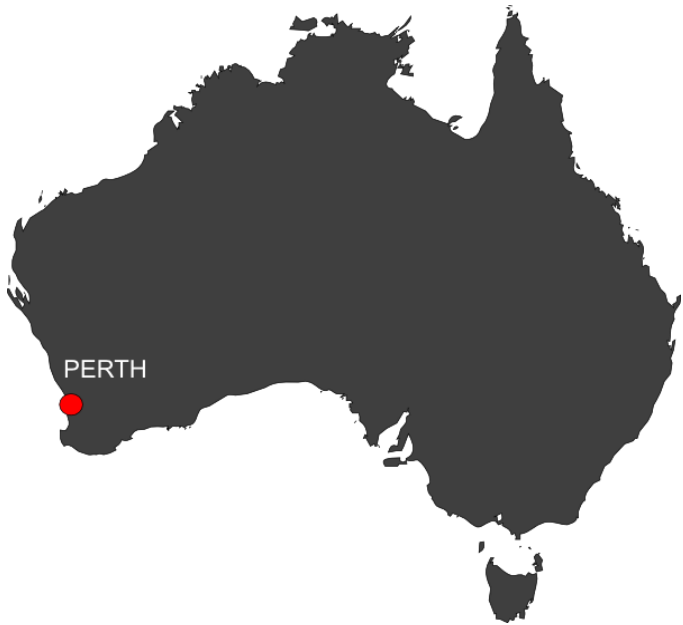

*Figure 2  Location of Perth, (Atlas, 2025)*

The dataset used was sourced from the NSW Government Open Data portal (NSW Governemnt, 2021). While the platform is an official and trustworthy government source, the dataset itself id not openly downloadable and requires a formal data request. This controlled access supports data integrity and privacy, adding to its reliability. This dataset includes hourly measurements of electricity and gas usage across various facility components. To support visualisation and pattern recognition, SAS Visual Analytics was employed to generate bar charts, parallel coordinate plots, and correlation matrices that highlight significant relationships among variables.

As shown in Figure 3figure 2, the correlation analysis of the facility's energy data highlighted key relationships among energy sources. Gas and Heating, show a near perfect correlation (0.98), indicating that gas usage is primarily driven by heating needs. Similarly, Electricity correlates strongly with Interior Lights (0.95) and Interior Equipment (0.94), suggesting that lighting and equipment are the main drivers of electricity demand. Interior Equipment and Interior Lights also show a strong relationship (0.92), implying these may be used during similar periods.

This analysis implies that reducing gas use for heating and improving lighting and equipment energy efficiency could effectively reduce overall energy demand. These findings suggest focusing on heating optimisation and energy-efficient lighting and equipment scheduling to manage both electricity and gas usage efficiently.

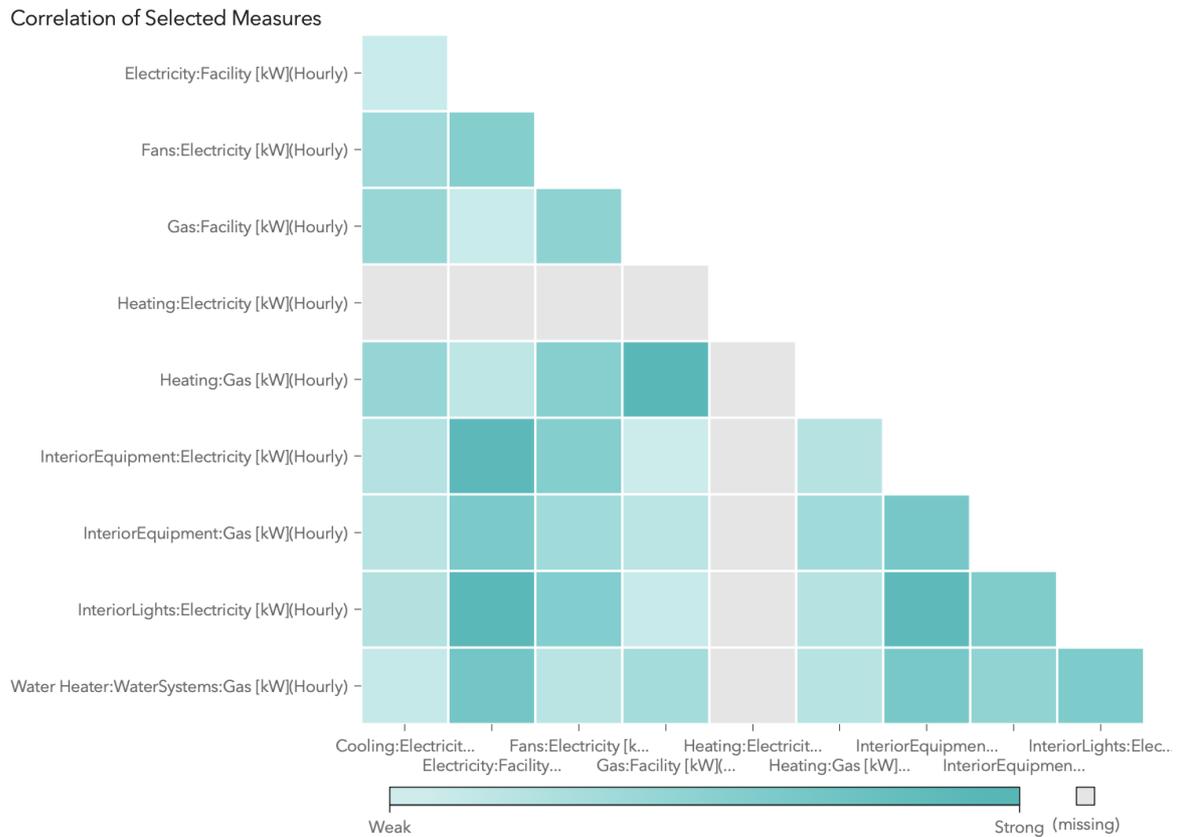

*Figure 3 Correlation Analysis*

Through the bar chart in Figure 3, and the parallel coordinate graphs in Figures 5-64-5, we can see that Electricity is strongly influences by Interior Lights (Importance of 1.000) and Interior Equipment (Importance of 0.946), with these two variables showing the highest correlations. This indicates that changes in lighting and equipment usage significantly drive the total electricity demand within the facility. The data shows that higher levels of interior lighting usage, particularly around the commonly observed values of 243.26 kW, are associated with higher overall electricity consumption. Meanwhile, interior equipment usage also consistently contributes to the facility's power demands, suggesting it represents a steady operational load. Fans and cooling also correlate with facility electricity usage, though to a lesser extent, highlighting their role in ventilation and environmental control. While cooling demand may vary seasonally, fan usage appears to be more consistent, likely driven by HVAC system requirements. This analysis suggests that focusing on lighting and equipment optimisation, possibly through automated control systems or operational scheduling, could be effective strategies to reduce total electricity demand without compromising functionality.

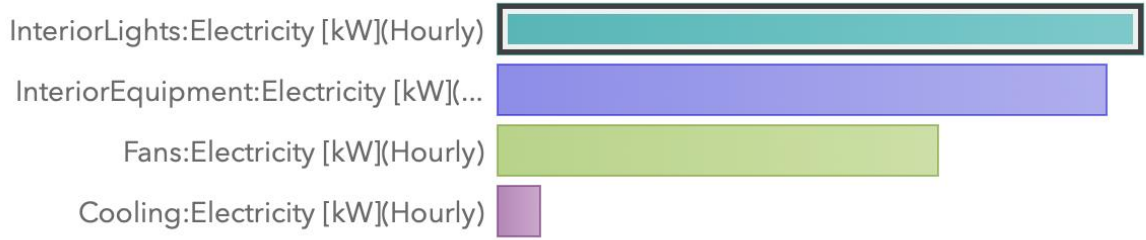

*Figure 4 Top factors affecting facility electricity use: lighting, equipment, fans, and cooling*

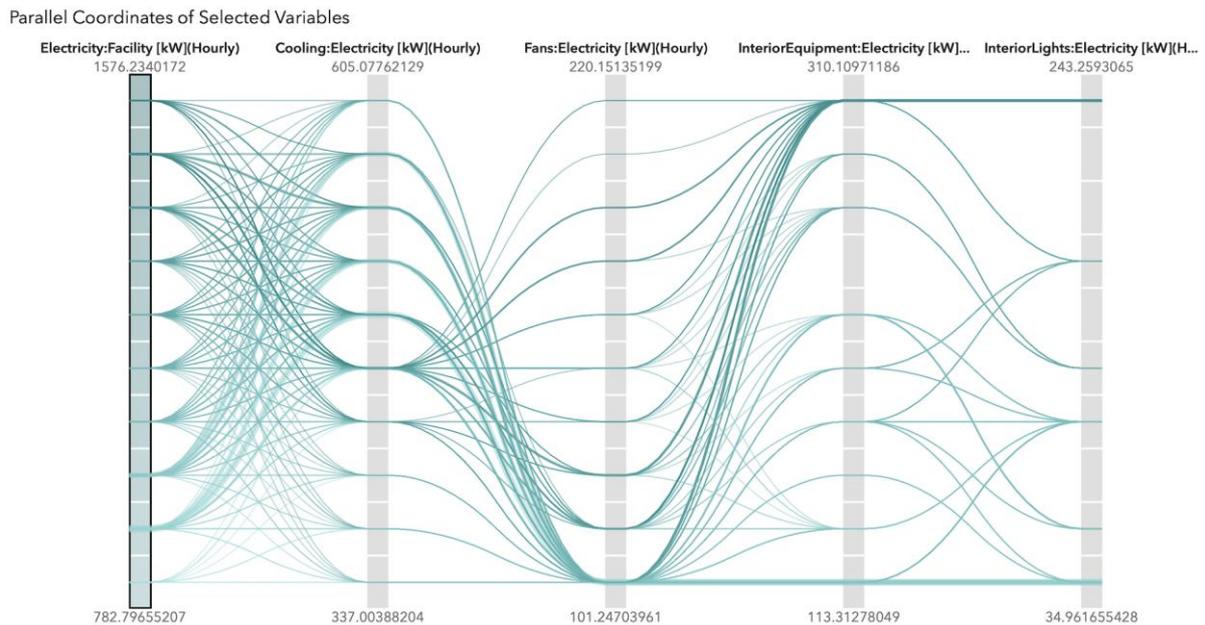

*Figure 5 Parallel Coordinate Graph*

This parallel coordinate plot in figure 5, shows the gas consumption patterns for the facility across three primary sources: Heating, Interior Equipment, and Water Heater. Heating dominates gas consumption, indicating that space heating likely represents the bulk gas usage. Interior Equipment and Water Heater have much lower and more consistent usage levels, suggesting they contribute minimally to overall gas demand. The similarity between Gas Facility and the Heating, also supports the fact that majority of the gas is used by Heating. Efficiency improvements targeting Heating could have a significant impact on reducing overall gas demand, while adjustments in Interior Equipment and Water Heater would likely yield smaller savings.

This analysis, based on data visualised with SAS Visual Analytics, reveals distinct patterns in energy usage within the hospital. The high correlation between gas and heating, and between electricity, lighting, and equipment, suggests targeted areas for energy optimisation. This methodology provides a foundation for developing strategies to manage electricity and gas usage more efficiently, which could lead to cost savings and reduced environmental impact for healthcare facilities.

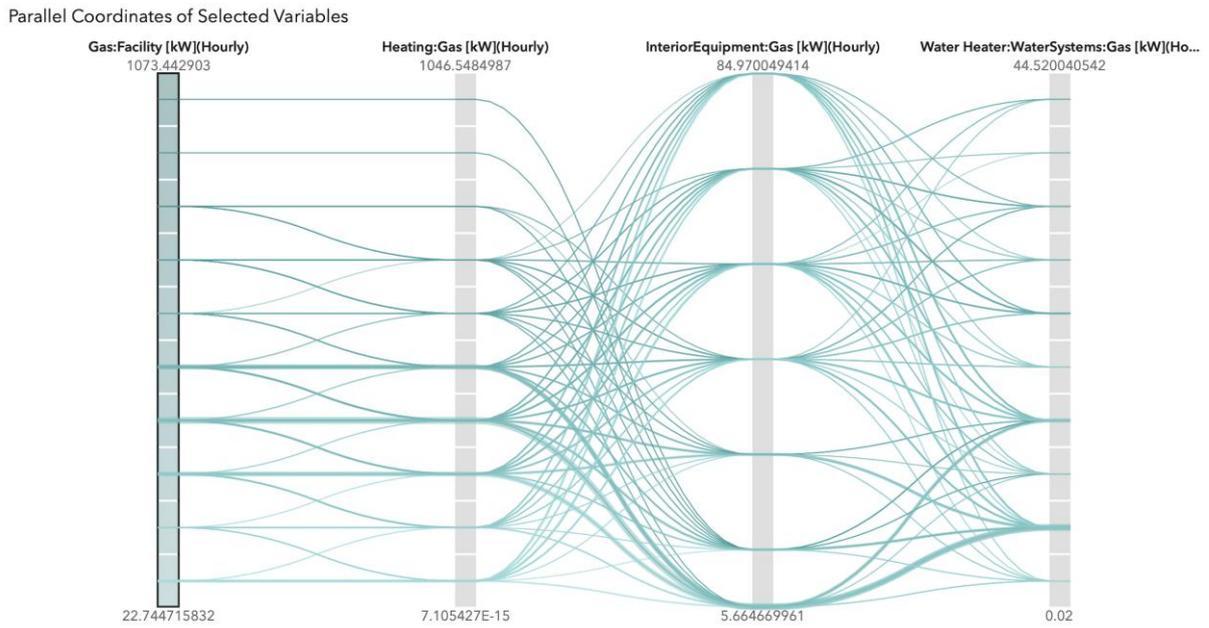

*Figure 6 Parallel Coordinate Plot of Selected Variables*

## 3.2 Implementation

In this research study, a framework for energy demand forecasting and load balancing in healthcare facilities is developed by integrating three advanced forecasting models—ARIMA, Prophet, and LSTM—along with genetic algorithm (GA) for load balancing. The methodology followed throughout the study is structured into distinct phases: model development, integration, and evaluation.

### 3.32.1 Model Development and Integration

Three different forecasting models were used in this study: ARIMA, Prophet, and LSTM, each chosen for their unique strengths in time series analysis.

#### 3.32.1.1 ARIMA Model (Autoregressive Integrated Moving Average)
The ARIMA model was implemented using Python's statsmodels library to capture linear trends in historical energy consumption data. The dataset, which contained electricity usage at hourly intervals, was pre-processed by converting timestamps into a datetime format, forward-filling missing values to maintain data continuity, and setting the 'Date/Time' column as the index for time series analysis.

The model was configured with the following hyperparameters:
- **p (autoregressive order):** 2
- **d (differencing order):** 1
- **q (moving average order):** 2

The ARIMA model was trained on the pre-processed time series data to predict future electricity demand over a 48-hour forecast horizon. The model's performance was evaluated using Mean Absolute Error (MAE) to measure the average magnitude of errors in predictions and Root Mean Squared Error (RMSE) to emphasise larger errors, highlighting extreme mismatches between actual and predicted values. Algorithm 1 presents the ARIMA forecasting with GA-based load optimization.

*Algorithm 1 Pseudocode: ARIMA Forecasting with GA-based Load Optimization*

```
Begin
// Step 1: Forecast Future Load using ARIMA
Load and preprocess time series energy demand data
Fit ARIMA model to the historical data
Set forecast_steps = 48  // Number of hours to forecast
Generate 'forecast' for the next 48 time steps using the fitted ARIMA model
// Step 2: Define GA Components
Define FitnessFunction(solution, forecast):
    Calculate absolute error between solution and forecast
    Return negative of the sum (to maximize fitness via minimization)
Define InitializePopulation(size, forecast):
    For each individual in population:
        Generate values between 0.8×forecast and 1.2×forecast
    Return population matrix
Define TournamentSelection(population, fitness, k):
    Randomly select k individuals from population
    Return the one with the highest fitness
Define SBX_Crossover(parent1, parent2, eta):
    For each gene:
        Generate random number
        Compute SBX blending coefficient 'beta'
        Generate two offspring based on SBX equations
    Return offspring1 and offspring2
Define AdaptiveMutate(offspring, generation, max_generations, base_prob, std_dev):
    Compute mutation probability as: base_prob × (1 - generation / max_generations)
    For each gene in offspring:
        With calculated probability, add Gaussian noise with std deviation = std_dev
    Return mutated offspring
// Step 3: Main GA Optimization Loop
Set population_size = 20
Set generations = 100
Initialize population using InitializePopulation
For generation = 1 to generations:
    Compute fitness for all individuals in the population using FitnessFunction
    Select two parents using TournamentSelection
    Apply SBX_Crossover to generate two offspring
    Apply AdaptiveMutate to each offspring
    Replace two individuals with the lowest fitness in the population using offspring
After all generations, select best_solution from population with highest fitness
End
```

### 3.3.2.1.2 Prophet Model

The Prophet model, developed by Facebook, was utilised for its ability to handle seasonal patterns, holiday effects, and trend changes, making it particularly suitable for healthcare energy forecasting. The dataset was reformatted to match Prophet's expected structure by renaming columns to 'ds' (date) and 'y' (energy consumption). Outliers (values beyond three standard deviations from the mean) were removed to prevent extreme data points from skewing the forecast.

The Prophet model's changepoint prior scale (which adjusts the flexibility of trend changes) and seasonality prior scale (which controls the strength of seasonal patterns) were optimised using GA. The GA iteratively refined parameter sets over 50 generations to minimise the RMSE of Prophet's predictions. Algorithm 2 presents the Prophet forecasting with GA-based load optimization.

*Algorithm 2  Pseudocode: Prophet Forecasting with GA-based Load Optimization*

```
Begin
// Step 1: Forecast Energy Demand using Prophet
Load historical energy demand into dataframe
Initialize Prophet forecasting model
Fit the model to the historical data
Create a future time frame for next 30 days (hourly intervals)
Generate forecast for future energy demand using the fitted model
Plot the forecasted demand
Select the first 50 hours from the forecast
Extract:
   - forecasted_demand (predicted values)
   - time_steps (corresponding timestamps)
// Step 2: Define Genetic Algorithm Components
Define FitnessFunction(solution, forecasted_demand):
   Compute sum of absolute differences between solution and forecasted_demand
   Return negative value (to minimize error)
Define InitializePopulation(pop_size, forecasted_demand):
   For each individual:
      Generate a vector with values randomly sampled between 0.8×forecast and 1.2×forecast
   Return the initialized population
Define AdaptiveMutate(offspring, current_generation, total_generations, base_prob, mean, std_dev):
   Compute mutation probability: base_prob × (1 - current_generation / total_generations)
   For each gene in offspring:
      With mutation probability:
         Add Gaussian noise (mean, std_dev)
   Return mutated offspring
// Step 3: Run Genetic Algorithm
Set population_size = 20
Set total_generations = 100

nitialize population using InitializePopulation
For generation = 1 to total_generations:
   Evaluate fitness of each individual using FitnessFunction
   Identify the worst individual (lowest fitness)
   Apply AdaptiveMutate to this individual
   Replace it in the population with the mutated version
Select best_solution from population with highest fitness
End
```

### 3.32.1.3 LSTM Model (Long Short-Term Memory)

The LSTM model was selected for its ability to capture non-linear relationships and long-term dependencies in time series data. The implementation was conducted using TensorFlow and Keras. The dataset was scaled using MinMaxScaler to normalise energy consumption values between 0 and 1, facilitating faster convergence during training. Time series sequences were generated using a sliding window approach, where the past 24 hours of energy usage were used to predict the next hour. The choice of a 24 hour lag was based on domain knowledge and an autocorrelation plot, which showed strong correlations between current energy consumption and the previous 24 hours. This aligns with daily operational cycles in healthcare facilities, where energy usage patterns typically repeat every 24 hours due to staff shifts, equipment usage, and patient intake cycles.

The LSTM model was structured with:
- Two LSTM layers: The first layer returned sequences for stacking, while the second produced the final output.
- Dropout layers: Inserted after each LSTM layer to prevent overfitting.
- Dense layers: Applied to produce the final prediction.

After training for 50 epochs, the model's performance was assessed using MAE and RMSE. Algorithm 3 presents the LSTM forecasting with GA-based load optimization.

*Algorithm 3 Pseudocode: Hyperparameter Tuning of LSTM using Genetic Algorithm*

```
Begin
// Step 1: Define Fitness Function for GA
Define FitnessFunction(solution):
    Extract hyperparameters: units1, units2, dropout1, dropout2 from solution
    Build LSTM model with:
       - LSTM layer with 'units1' neurons and return_sequences=True
       - Dropout layer with rate = dropout1 / 10
       - LSTM layer with 'units2' neurons and return_sequences=False
       - Dropout layer with rate = dropout2 / 10
       - Dense layer with 25 neurons
       - Final Dense output layer with 1 neuron
    Compile model using Adam optimizer and mean squared error loss
    Train model on training data for 5 epochs
    Predict on test data
    Compute and return negative of Mean Absolute Error (MAE) between predicted and actual values (in original scale)
// Step 2: Set GA Parameters
Initialize GeneticAlgorithm with:
    - Number of generations = 5
    - Number of parents mating = 3
    - Population size = 5
    - Number of genes = 4
    - Gene space:
        • units1: 20 to 100
        • units2: 20 to 100
        • dropout1: 1 to 5
        • dropout2: 1 to 5
    - Fitness function = FitnessFunction
// Step 3: Run GA Optimization
Run the Genetic Algorithm
Retrieve best_solution and its corresponding best_fitness
// Step 4: Build Optimized LSTM Model
Extract best hyperparameters from best_solution
Construct optimized LSTM model with the selected units and dropout rates
Compile model using Adam optimizer and mean squared error loss
// Step 5: Train Optimized Model
Train the optimized model on training data:
    - Number of epochs = 50
    - Batch size = 32
```

```
      - Validate on test data
   // Step 6: Evaluate the Model
   Make predictions on test data
   Apply inverse scaling to both predictions and actual test data
   Calculate evaluation metrics:
      - Mean Absolute Error (MAE)
      - Root Mean Squared Error (RMSE)
   Print the final MAE and RMSE values
   End
```

### 3.32.1.4 Integration with genetic algorithms for Load Balancing

In addition to forecasting, a GA was employed to optimize load balancing, ensuring that energy demand was distributed as efficiently as possible across hospital resources. The core of this approach lay in a fitness function specifically designed to minimize the absolute difference between forecasted demand and proposed load allocations. To seed the search, the initial population was generated with random values within 20 percent of the forecasted demand, after which tournament selection was used to choose the fittest individuals for crossover. A blend crossover operator then combined parent solutions to produce new offspring, and Gaussian-noise mutation was applied to introduce variability and encourage exploration of the solution space. After running the algorithm for 100 generations, the optimized load allocation was compared directly with the forecasted demand, demonstrating the GA's ability to substantially reduce imbalances and improve overall system efficiency.In addition to forecasting, a GA was employed to optimise load balancing, ensuring efficient distribution of energy demand across hospital resources.

The proposed approach includes a fitness function aimed at minimizing the absolute difference between forecasted demand and proposed load allocations. Population initialization involved generating random values within 20% of the forecasted demand. Tournament selection was applied to select parents for crossover, where blend crossover was utilized to combine parent values and generate new solutions. To encourage exploration, Gaussian noise was introduced during the mutation process, promoting the discovery of diverse solutions.

### 3.3.5 SHAP Analysis for Model Interpretability

To interpret the forecasting models and understand the contribution of different input features, SHAP analysis was performed. Since ARIMA and Prophet models do not inherently support SHAP interpretation, surrogate models were employed to approximate their predictions, and LSTM's predictions were analysed using an XGBoost surrogate model.

### 3.3.5.1 SHAP Analysis for ARIMA

For ARIMA, lagged features were created by incorporating the past 24 hourly values as input predictors. These lagged values help capture the temporal dependencies in the data. After preprocessing the dataset with these features, a RandomForestRegressor was trained on them to approximate the ARIMA model's predictions. This surrogate model enabled SHAP analysis, which calculates the contribution of each feature to the model's output. The process has been presented in Figure 7. The process for SHAP analysis in the ARIMA model begins with the creation of lagged features, incorporating the past 24 hourly values into the dataset to capture temporal dependencies. The dataset is then split into training and testing sets to facilitate model evaluation. A RandomForestRegressor is subsequently trained to predict energy consumption based on these lagged features, serving as a surrogate model to approximate ARIMA's predictions. Finally, SHAP values are computed to interpret the contribution of each feature to the model's output, providing insights into feature importance and model behavior.

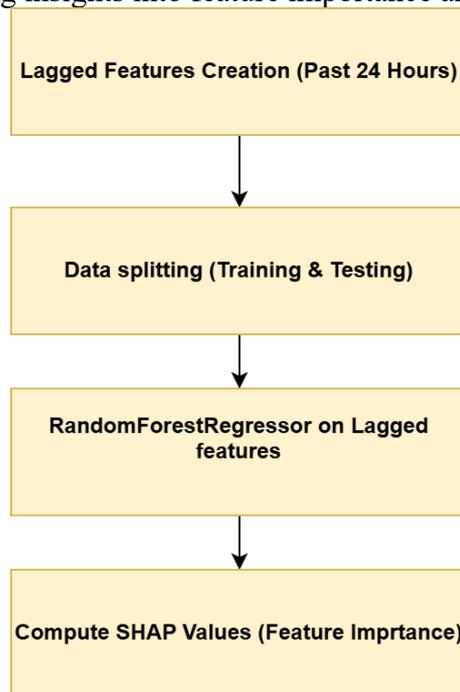

*Figure 7 SHAP Analysis For ARIMA - Flowchart*

**3.3.5.2** SHAP Analysis for Prophet

For the Prophet model, its forecasting capabilities were decomposed into trend and seasonal components, which are crucial for understanding how it predicts energy demand. Like with ARIMA, a RandomForestRegressor was used as a surrogate model to approximate Prophet's predictions, allowing SHAP analysis (Figure 8). The process for SHAP analysis in the Prophet model involves decomposing the forecast into trend, weekly, and yearly components to capture distinct temporal patterns. Lagged features representing the past 24 hours are then generated and incorporated into a surrogate model. A RandomForestRegressor is trained using these lagged features and Prophet's components to approximate the model's predictions. Subsequently, SHAP values are computed to assess the contribution of each feature and component, providing a comprehensive interpretation of feature importance and model behavior.

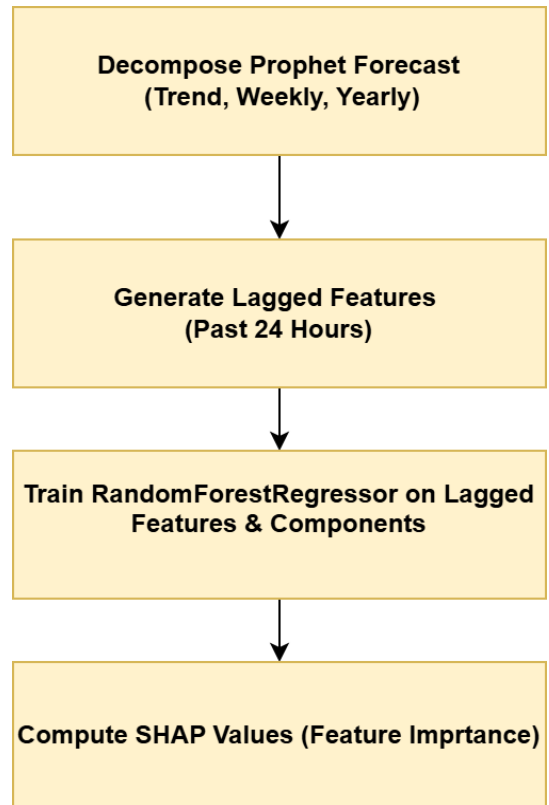

*Figure 8 SHAP Analysis For ARIMA - Flowchart*

### 3.3.5.3 SHAP Analysis for LSTM

The LSTM model, being a deep learning approach, directly generates predictions based on past time steps. However, to enable SHAP analysis, a XGBoost surrogate model was used to approximate the LSTM's predictions, making it possible to interpret the feature importance. The SHAP analysis process for the LSTM model begins with the generation of lagged features for the past 24 hours, serving as input predictors. A XGBoost model is then trained using these lagged features to approximate the LSTM model's predictions (Figure 9). Finally, SHAP values are calculated through the surrogate model to evaluate the influence of each feature on the LSTM model's outputs, enabling a comprehensive interpretation of feature importance and model behavior.

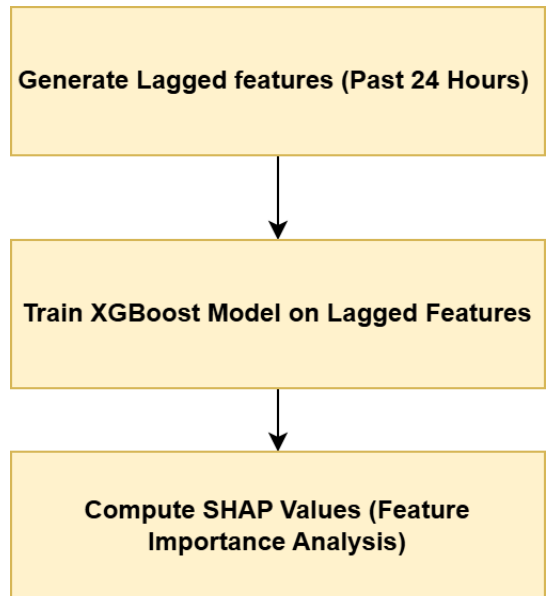
*Figure 9 SHAP Analysis For ARIMA - Flowchart*

## 3.4 Evaluation

The performance of the forecasting and load balancing models was evaluated using multiple metrics. For the forecasting models (ARIMA, Prophet, and LSTM), accuracy was assessed using MAE and RMSE. These metrics helped quantify each model's predictive capability in replicating actual energy consumption patterns. For the load balancing optimisation using GA, the evaluation focused on the efficiency of the energy distribution strategy. Key evaluation metrics included energy savings, peak demand reduction, and cost minimisation. Additionally, the GA's adaptability to changing energy demands (as predicted by the forecasting models) was analysed by comparing energy consumption under the optimised load balancing strategy with baseline consumption levels.

In summary, this approach not only focused on the prediction of energy demand but also addressed the critical aspect of efficient energy usage in healthcare facilities, ensuring that the system can handle dynamic demand while maintaining optimal performance.

1. **Results**

The following sections provide the evaluation metrics for the models.
### 4.1 Evaluation

The performance of the ARIMA, Prophet, and LSTM models was evaluated using Mean Absolute Error (MAE) and Root Mean Squared Error (RMSE), two widely used metrics for assessing forecasting accuracy. Lower values of these metrics indicate better predictive performance. The results, summarised in Table 3, highlight significant differences in accuracy among the models.

The ARIMA model exhibited the highest error rates, with an MAE of 87.73 and an RMSE of 125.22. This indicates ARIMA's limitations in capturing the complex, non-linear fluctuations in energy demand, leading to significant deviations between predicted and actual values. Despite being a strong statistical approach, ARIMA struggled with the dynamic nature of energy consumption patterns, resulting in the highest forecast errors.

The Prophet model performed notably better than ARIMA, achieving an MAE of 59.78 and an RMSE of 81.22. This improvement can be attributed to Prophet's ability to model seasonality and trend changes more effectively. However, while Prophet reduced forecasting errors compared to ARIMA, it still exhibited challenges in capturing short-term demand variations, leading to residual discrepancies between forecasted and actual demand. The LSTM model demonstrated the best performance, with a significantly lower MAE of 21.69 and RMSE of 29.96. This highlights LSTM's ability to learn complex, non-linear dependencies in energy demand data. By leveraging past consumption trends and capturing both short- and long-term patterns, LSTM provided the most accurate forecasts, with minimal deviation from actual energy demand.

As visualised in Figure 10, the results indicate that while ARIMA and Prophet have their strengths in time series forecasting, they fall short in dynamic and high-variability environments like healthcare energy demand. The LSTM model, benefiting from deep learning capabilities and GA-based hyperparameter optimisation, proved to be the most effective, achieving the lowest error rates and delivering more reliable predictions.

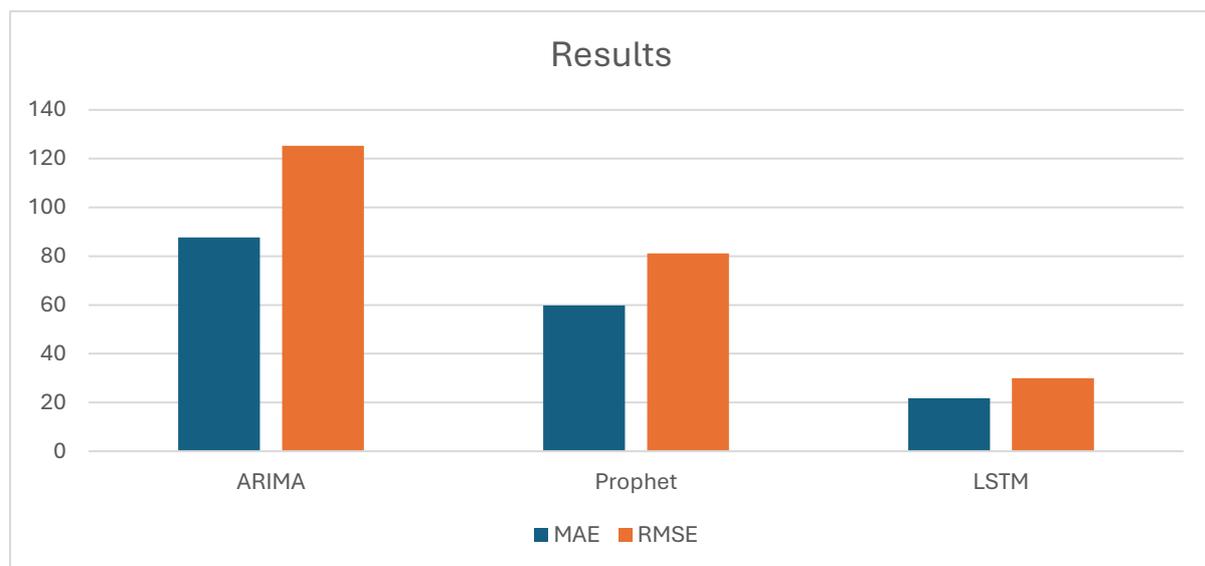

*Figure 10 Comparison of forecasting models based on MAE and RMSE metrics*

## 4.2 Graph Analysis

The graphical representation of the forecasting results further highlights the strengths and limitations of each model. The ARIMA model's forecasted demand is depicted as a relatively stable blue line in Figure 11, while the optimised load allocation exhibits significant fluctuations. This discrepancy suggests ARIMA's difficulty in capturing the complexity of real-world energy demand variations. The high degree of variation in load allocation indicates that ARIMA's linear assumptions fail to account for the non-linear patterns inherent in energy consumption, leading to higher forecasting errors. The model struggles particularly with capturing short-term demand spikes, which are crucial for efficient energy distribution in healthcare facilities.

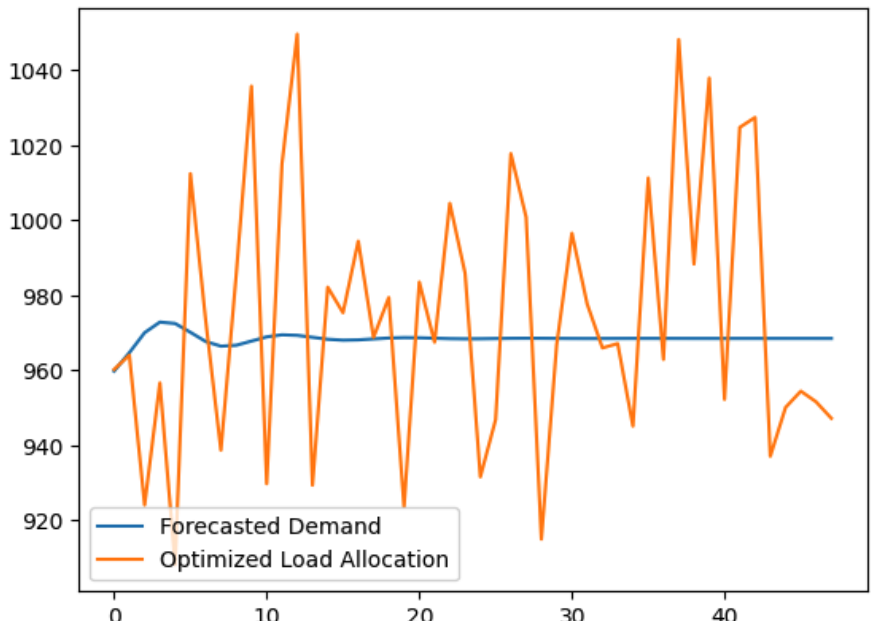

*Figure 11 ARIMA: Forecasted Demand v Optimised Load Allocation*

In contrast, the Prophet model demonstrates improved forecasting accuracy, as seen in Figure 12. The red dashed line, representing Prophet's predictions, follows a smoother trajectory and better captures overall demand trends. However, despite its ability to model seasonality and trend changes effectively, the green line indicating optimised load allocation still exhibits fluctuations. This suggests that Prophet, while superior to ARIMA in identifying long-term demand patterns, does not fully capture the rapid short-term variations necessary for precise energy forecasting.

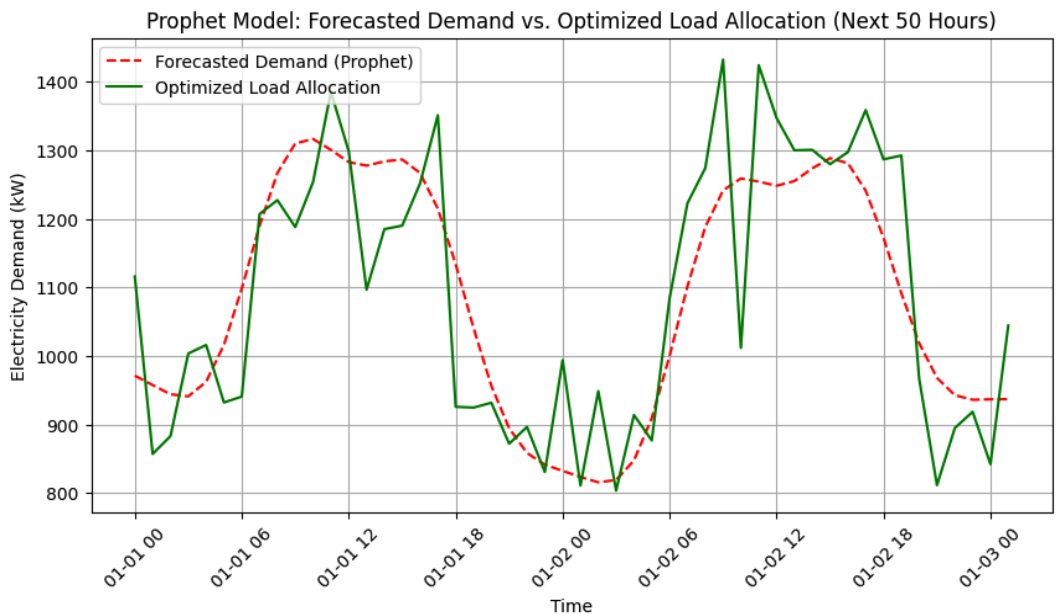

*Figure 12 Prophet: Forecasted Demand v Optimised Load Allocation*

The LSTM model, as shown in Figure 13, provides the closest alignment between forecasted and actual demand. The orange dashed line (LSTM predictions) closely follows the blue line (actual demand), highlighting its ability to adapt to non-linear demand variations. The minimal deviation between these lines indicates that LSTM effectively captures both short-term fluctuations and long-term dependencies, making it the most reliable model for energy demand forecasting. The model's deep learning architecture enables it to recognise intricate patterns in the data, significantly reducing forecasting errors compared to ARIMA and Prophet.

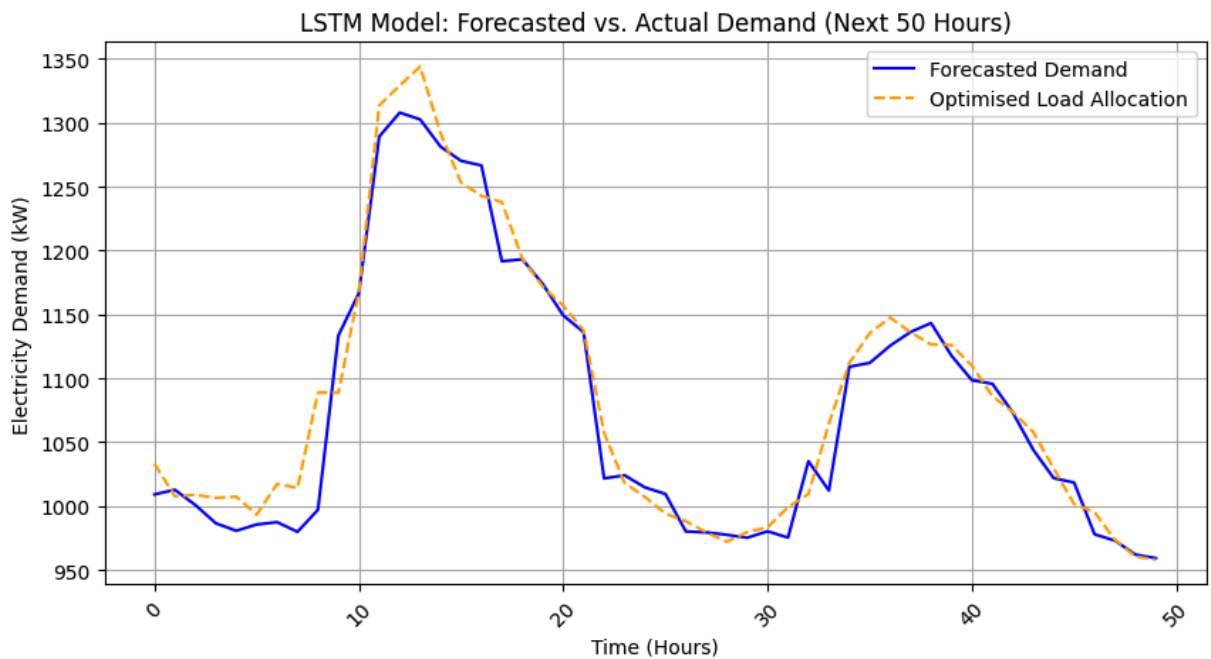

*Figure13 LSTM: Forecasted Demand v Optimised Load Allocation*

## 4.3 SHAP Analysis

To further understand the impact of different input features on each model's predictions, SHAP values were computed (Figures 14-15).. These values quantify the contribution of each feature to the model's predictions.

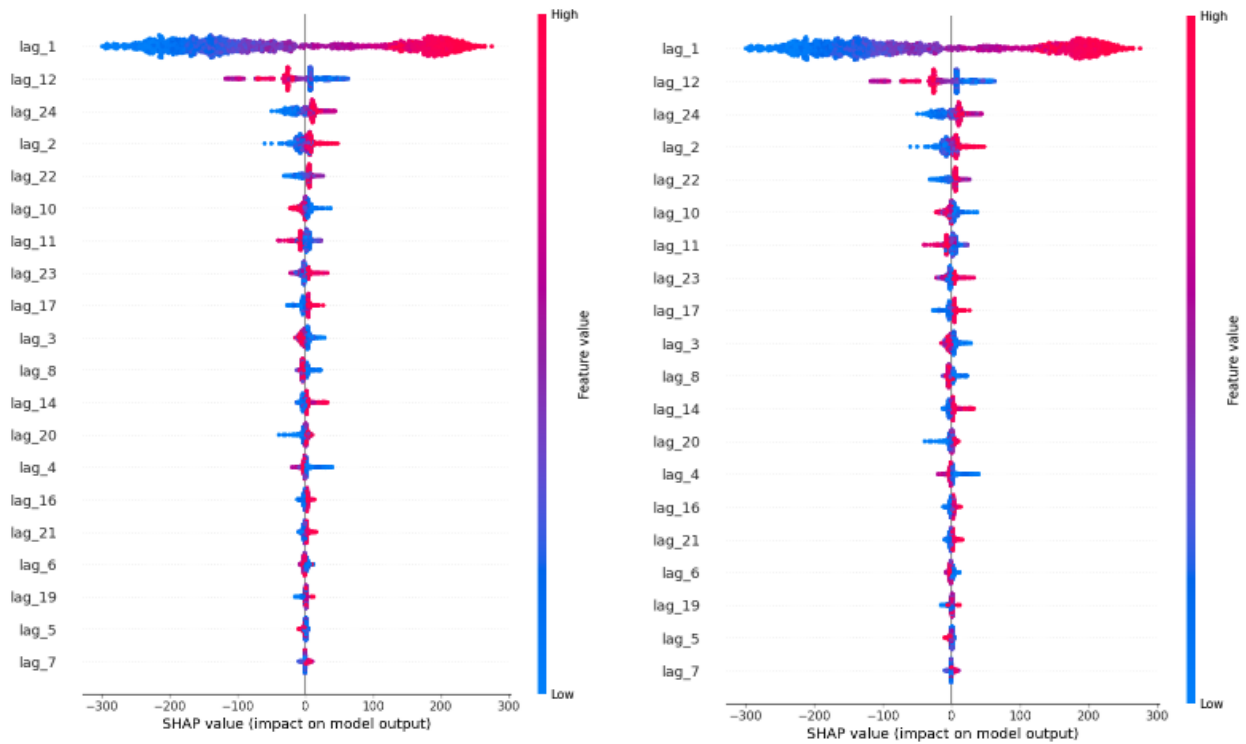

*Figure 14 SHAP Analysis of ARIMA (Left), and Prophet (Right)*

For the ARIMA model, the most influential feature was lag_1, with a mean SHAP value of 153.89, indicating that the previous time step had the highest impact on predictions. Other notable features included lag_12 (25.46) and lag_24(13.22), suggesting that historical data from 12 and 24 time steps prior also played a role. However, the lower SHAP values of other lags indicate that ARIMA primarily relies on a few past observations, limiting its ability to model complex patterns.

The Prophet model exhibited a similar reliance on lag_1 (153.89), lag_12 (25.46), and lag_24 (13.22), as shown in Figure 10. This highlights Prophet's tendency to use recent and periodic trends in forecasting. However, the moderate contribution of longer lags suggests Prophet considers broader trends but does not significantly improve on ARIMA in terms of feature importance distribution.

In contrast, the LSTM, Figure 15, model demonstrated a more balanced distribution of feature importance. lag_1 remained the most critical feature (163.85), but lag_12 (30.04) and lag_2 (13.44) also had substantial contributions. Unlike ARIMA and Prophet, LSTM leveraged a wider range of past observations, including mid-range lags like lag_7 (6.79) and lag_8(7.11), allowing it to capture both short-term and long-term dependencies effectively. The relatively even spread of SHAP values across multiple features highlights LSTM's superior capacity for learning complex temporal relationships.

These SHAP results reinforce the model evaluation findings, demonstrating that ARIMA and Prophet rely heavily on a limited set of past observations, whereas LSTM effectively integrates information from various time steps, making it the most robust model for energy demand forecasting.

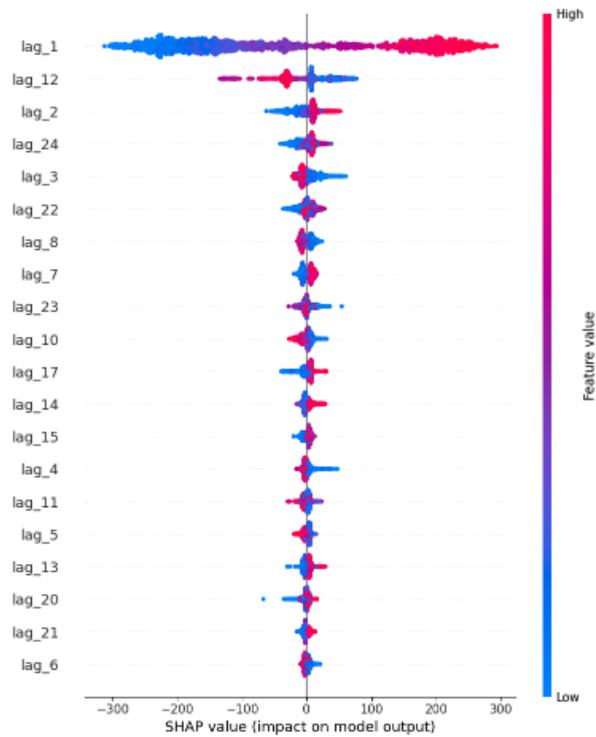

*Figure 15 SHAP Analysis of LSTM*

## 4.4 Model Performance

The ARIMA model, optimised with hyperparameters (p=2, d=1, q=2), exhibited the highest error rates, with an MAE of 87.73 and an RMSE of 125.22. Despite effectively capturing linear trends, ARIMA's reliance on these assumptions limited its ability to model the non-linear and highly dynamic nature of energy demand. This resulted in significant deviations between predicted and actual values, particularly during short-term fluctuations. The model struggled to align with real-world demand variations, leading to significant forecasting errors.

The Prophet model demonstrated improved accuracy compared to ARIMA, achieving an MAE of 59.78 and an RMSE of 81.22. Prophet's ability to model seasonality and trend changes contributed to better alignment with long-term demand patterns. However, the model still faced challenges in capturing rapid short-term variations in energy consumption, leading to residual forecast inaccuracies. The graphical analysis of its forecasts showed that while it captured overall trends more effectively than ARIMA, it still failed to fully match the dynamic fluctuations in energy demand.

The LSTM model significantly outperformed both ARIMA and Prophet, achieving an MAE of 21.69 and an RMSE of 29.96. By leveraging a deep learning architecture capable of recognising complex temporal dependencies, LSTM effectively modelled both short-term demand fluctuations and long-term trends. The model's deep learning capabilities allowed it to adapt to real-time energy variations, making it the most reliable forecasting model in this study. The graph analysis showed that LSTM predictions closely followed actual demand patterns, demonstrating its ability to capture non-linear relationships more effectively than the other models.

To further interpret the models' decision-making processes, SHAP analysis was conducted. The results revealed that both ARIMA and Prophet primarily relied on recent time steps, particularly lag_1, with secondary influence from lag_12 and lag_24. This suggests that both models placed heavy weight on immediate past observations but failed to integrate a broader range of historical data effectively. In contrast, LSTM exhibited a more balanced distribution of feature importance, leveraging mid-range and long-range dependencies. This ability to integrate multiple time steps contributed to its superior forecasting accuracy.

Overall, these results emphasise the advantages of deep learning approaches for energy demand forecasting in dynamic environments such as healthcare facilities. While ARIMA and Prophet provide reasonable forecasts under structured conditions, their reliance on limited past observations and difficulty capturing non-linear trends reduce their effectiveness. LSTM's ability to adapt to non-linear patterns, capture both short- and long-term dependencies, and optimise predictive accuracy makes it the most effective model for this application.

## 2. Discussion

The results of this research demonstrate the varying effectiveness of ARIMA, Prophet, and LSTM models in predicting energy demand in healthcare facilities. Among the models, LSTM exhibited the highest accuracy, with the lowest MAE (21.69) and RMSE (29.96). This highlights LSTM's ability to capture both short- and long-term dependencies in energy use while effectively modelling complex, non-linear relationships in the data. The graphical analysis reinforced these findings, as LSTM's forecasted demand closely aligned with actual values, minimising deviations and ensuring reliable energy predictions.

Conversely, ARIMA, despite its well-established statistical framework, struggled with the unpredictable nature of energy demand in healthcare settings. Its reliance on linear assumptions resulted in significant forecasting errors, with an MAE of 87.73 and RMSE of 125.22. The model's limited capacity to adapt to sudden demand spikes made it less suitable for dynamic environments. Prophet offered moderate improvement, achieving an MAE of 59.78 and RMSE of 81.22. While Prophet effectively captured seasonality and long-term trends, its inability to model rapid short-term fluctuations limited its forecasting precision. SHAP analysis further confirmed these trends, showing that ARIMA and Prophet primarily relied on recent time steps, whereas LSTM integrated a broader range of historical data, leading to superior forecasting performance.

The integration of GA added a crucial optimisation layer, refining model efficiency and improving load balancing. By tuning hyperparameters—such as Prophet's changepoint prior scale and LSTM's dropout rates—GA enhanced predictive performance while optimising energy distribution. Over 100 generations, GA successfully minimised demand imbalances, ensuring adaptive and efficient energy allocation. The optimised load allocation results, particularly when coupled with LSTM's accurate forecasts, demonstrated a significant reduction in fluctuations, highlighting the effectiveness of AI-driven load management strategies.

This research successfully achieved its primary objective: evaluating and comparing forecasting models while integrating GA for energy optimisation. The assessment of predictive accuracy through MAE and RMSE confirmed LSTM's superiority over ARIMA and Prophet. Additionally, the iterative refinement of GA-based load balancing strategies met the objective of optimising energy distribution. The combination of LSTM and GA emerged as the most effective solution, balancing forecasting precision with real-time adaptability.

Furthermore, the research effectively addressed key questions related to AI-driven energy forecasting. LSTM consistently outperformed ARIMA and Prophet, providing the most accurate demand predictions. GA significantly enhanced load balancing, ensuring efficient energy management. The LSTM-GA framework stood out as a practical and scalable solution, balancing accuracy, efficiency, and cost-effectiveness. Finally, the study provided actionable recommendations for real-world implementation, such as integrating real-time data updates, hybrid modelling approaches, and cost-aware energy management strategies, contributing to improved sustainability and operational efficiency in healthcare facilities.

## 3. Conclusion and Future Work

This research successfully developed and evaluated a framework for energy demand forecasting and load balancing in healthcare facilities by integrating three predictive models— ARIMA, Prophet, and LSTM—along with a GA for optimisation. Through comprehensive model development and evaluation, the study provided critical insights into the comparative performance of these forecasting techniques and their potential for enhancing energy efficiency in dynamic environments.

The results demonstrated that while ARIMA and Prophet were effective in capturing linear trends and seasonal variations, respectively, LSTM significantly outperformed them in terms of predictive accuracy, achieving the lowest MAE and RMSE values. The graphical and SHAP analyses further confirmed that LSTM's ability to leverage both short- and long-term dependencies allowed it to model complex energy consumption patterns more effectively. Additionally, the integration of GA successfully optimised both forecasting model parameters and load balancing decisions, enhancing the system's adaptability to fluctuating energy demands. The optimised load allocation results highlighted the effectiveness of AI-driven strategies in minimising demand imbalances and improving energy distribution efficiency.

The findings of this study underscore the potential of AI-driven energy management strategies in healthcare settings, where accurate forecasting and efficient load allocation are critical for operational sustainability. The proposed approach not only improves energy utilisation but also contributes to cost reduction and environmental sustainability by optimising energy demand management.

While this research achieved its objectives, certain limitations remain. The reliance on historical data, rather than real-time implementation, limits the framework's ability to dynamically adjust to sudden changes in energy consumption. Future work could focus on deploying the framework in live healthcare environments to test its adaptability to real-time energy data. Exploring hybrid forecasting models—such as combining Prophet's seasonality strength with LSTM's ability to capture non-linear patterns—may further enhance accuracy.

Additionally, investigating alternative optimisation techniques, such as Particle Swarm Optimisation (PSO) or Reinforcement Learning (RL), could provide more efficient load balancing solutions. Expanding the application of this framework across multiple healthcare facilities with diverse energy consumption patterns would validate its scalability and robustness.

In conclusion, this study establishes a solid foundation for leveraging AI in energy demand forecasting and load balancing within healthcare facilities. By integrating deep learning with evolutionary optimisation techniques, the LSTM-GA framework emerged as a highly effective solution for improving energy efficiency. The results highlight the importance of intelligent energy management systems in reducing costs, supporting sustainability, and enhancing the reliability of power distribution in healthcare environments. These findings pave the way for future advancements in AI-driven energy optimisation, bridging the gap between predictive analytics and real-world energy management solutions.